# Grammatical facial expression recognition using customized deep neural network architecture

Devesh Walawalkar
dwalawal@andrew.cmu.in

## Abstract

This paper proposes to expand the visual understanding capacity of computers by helping it recognize human sign language more efficiently. This is carried out through recognition of facial expressions, which accompany the hand signs used in this language. This paper specially focuses on the popular Brazilian sign language (LIBRAS). While classifying different hand signs into their respective word meanings has already seen much literature dedicated to it, the emotions or intention with which the words are expressed haven't primarily been taken into consideration. As from our normal human experience, words expressed with different emotions or mood can have completely different meanings attached to it. Lending computers the ability of classifying these facial expressions, can help add another level of deep understanding of what the deaf person exactly wants to communicate. The proposed idea is implemented through a deep neural network having a customized architecture. This helps learning specific patterns in individual expressions much better as compared to a generic approach. With an overall accuracy of 98.04%, the implemented deep network performs excellently well and thus is fit to be used in any given practical scenario.

## Introduction

Sign language is an essential medium used by deaf people to communicate with other people in their environment. As sign language doesn't have a speech component through which an average human conveys the emotion behind what he says or does, the facial expressions assume this important role in a sign language.

A computer trained only to understand the language through hand gestures would fail to understand the semantic and structural level context of what the person tries to convey. A lot of literature on this topic [1,3,4,7,8,9,11,15,16,17] has primarily focused on sign language recognition through hand gestures only without considering its facial expressions aspect. Combining classification of facial expressions along with hand gestures would result in a more efficient interpretation [13,18].

These facial expressions are called `Grammatical Facial Expressions' (GFEs) as they help to resolve the semantic level ambiguity in human sign language. Facial expression recognition has attracted attention over recent years, because of the fact that it can be very useful in many applications such as speech recording systems which uses sign language to normal language text conversion, subtitling a video in which sign language is conveyed etc. Neural network techniques are used for this topic as it is very efficient in learning complex functions when given enough training data. Previous work on GFE classification [2] is based upon traditional classification methods, thus in turn failing to leverage potential of recent deep learning developments. This paper presents comparison of proposed model performances with those stated by Freitas et al. [9], with both models being computed on the same dataset. Performance comparisons with a generic fully connected neural network have also been presented.

The presented paper structure is as follows: 1] Demonstration of the fundamental classes (markers) through which wide variety of GFEs can be classified. 2] The incorporated dataset and its detailed description, 3] Implementation of customized deep neural network architecture, 4] Network initialization and its hyper parameter tuning, 5] Cost function and Optimization algorithm used, 6] Binary and Multiclass classification system performance results, 7] Comparison of binary classification performance with that of an accepted method present in literature 8] Final discussion of achieved results and its implications.

## Importance of Grammatical Facial Expressions

Sign language consists of mainly two components: manual and non-manual. The manual components consist of hand shape, palm orientation and arm movement. The non-manual components consist of facial expressions, pose and mouth movement. Some signs can be distinguished from manual components only, while rest need the additional non-manual component to distinguish them. The Brazilian sign language system consists of certain words which have nearly identical hand sign formation. They differ from each other only in terms of the facial expression with which they are said. Hence Sign language recognition only through manual cues leads to inefficient and ambiguous classification.

Facial expressions play a vital role in effectively communicating the information through to the listener. In a language such as English, the exclamation mark, question mark, the comma etc. provides the emotion attached to a said sentence by the source. As reshuffling the comma in different positions in same sentence can completely change the meaning of it, in the same way change or absence of GFEs in a sign language can completely change its meaning.

# Database Collection

This paper is based upon empirical results computed on `Grammatical Facial Expression Dataset', created by Freitas et al. [9] and obtained under public license from University of California, Irvine Machine learning repository [14]. This dataset is based upon facial expression made by a sign language performer (further mentioned as user) captured through individual video frames. There are eight fundamental types of grammatical markers in Brazilian sign language [Libras] system as stated by Brito [5] and de Quadros et al. [6]. These are as follows, along with their meanings:

- **Wh question**: generally used for questions with Who, What, When, Where, How and Why;
- **Yes/no question**: used when asking a question to which there is a `yes' or `no' answer;
- **Doubt question**: This is not a `true' question since an answer is not expected. However, it is used to emphasize the information that will be supplied;
- **Topic**: used when one of the sentence's constituents is displaced to the beginning of the sentence;
- **Negative**: It is used in negative sentences;
- **Assertion**: used when making assertions;
- **Conditional**: used in subordinate sentence to indicate a prerequisite to the main sentence;
- **Focus**: used to highlight new information into the speech pattern;

The dataset consists of 225 videos recorded in five different recording sessions carried out with the user. In each session, one performance of each sentence was recorded. User was asked to perform the sentence from each of the above type (with an additional Relative marker type, which is used at start of a clause in the sentence). Examples of these mentioned markers present individually in sentences in the common English language is given as follows [9]:

- **Conditional:**
  1] If you miss, you lose.
  2] If it's sunny, I go to the beach.
- **Assertion:**
  1] I bought that!
  2] I work there!
- **Negative:**
  1] I never have been in jail!
  2] I didn't do anything!
- **Relative:**
  1] That enterprise? ... Its business is technology!
  2] The girl who fell from bike? ... She is in the hospital!
- **Focus:**
  1] The bike is BROKEN.
  2] It was WAYNE who did that!
- **Topics:**
  1] I have a notebook!
  2] Fruits ... I like pineapple!
- **Doubt questions:**
  1] Did you GRADUATE?
  2] Did Wayne buy A CAR?
- **Wh-questions:**
  1] What is this?
  2] Where do you live?
- **Yes/no questions:**
  1] Did he go away?
  2] Is this yours?

Multiple frames were captured from each of these marker videos and predefined attribute face points (Figure 1) were located in each of these frames. The X, Y (frontal image plane) and Z (depth) coordinates of each of these 100 attributes, for each frame were recorded using a Microsoft Kinect $^{TM}$ sensor.

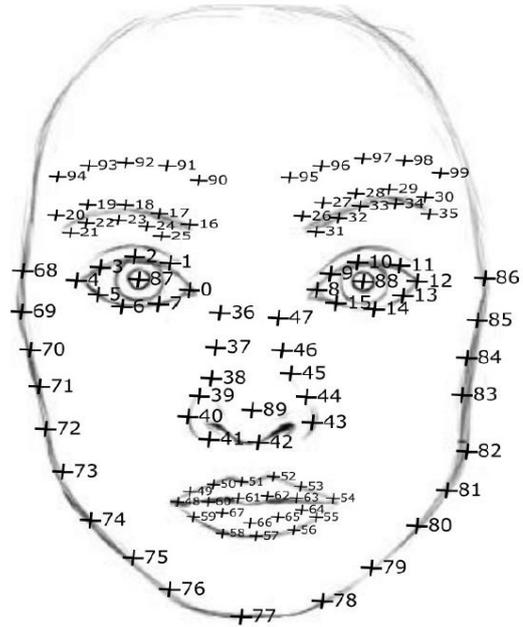

Fig. 1: Attribute point locations on User face

These frames were then hand classified as a binary classification task for each individual class with help of a sign language expert. This procedure was implemented for two users (A and B) so as to reduce any particular user bias present in acquired data. The dataset contains 27965 frames in total, classified into 18 different classes (9 for each user). Description of dataset constituents can be seen in Table 1.

Table 1: No. of positive and negative samples for each GFE class

| GFE class type | Positive samples | Negative samples |
|---|---|---|
| Assertion | 541 | 644 |
| Yes/no question | 734 | 841 |
| Negative | 568 | 596 |
| Topic | 510 | 1789 |
| Conditional | 448 | 1486 |
| Doubt question | 1100 | 421 |
| Focus | 446 | 863 |
| Relative | 981 | 1682 |
| Wh question | 643 | 962 |

The implemented test set comprises of 30% of the total dataset available. This gives a sample (frame) count of 400 - 450 samples per class (for binary classification) for each user.

## Data Pre-Processing

For each attribute, its (X, Y) coordinate points are given in pixels, whereas its Z coordinate is given in mm. As both units are different and hence their numerical ranges being different, Z score standardization is performed on dataset before using it in experimentation. This also makes the learnt model invariant to the location of face in captured frame (i.e. having a different set of attribute numerical values). Some isolated coordinate values missing from the dataset are represented by a placeholder value of '0.0'. Such random values could lead to wrong model learning. Hence, these values are replaced by mean of that particular attribute point's (either X, Y or Z) remaining sample values present in dataset. This particular modification was positively supported by enhanced model performance. Each of the marker's binary classification dataset contains an unequal number of positive and negative cases, with on average negative ones being much greater than positive ones (refer Table 1). This might lead the model to be slightly biased towards learning the negative pattern. Hence for training, equal number of both classes are considered. This does lead to an appreciable increase in model performance.

## Deep Neural Network Architecture

For this model, a customized feed-forward deep network architecture was implemented. It consists of two hidden layers along with the standard input and output layers. The entire customized architecture can be referred to in Figure 2. Here, for each sample (frame) the attribute point's standardized X, Y, Z coordinates are fed to a single neuron in the first hidden layer. Thus, 100 neurons present in first hidden layer are tuned to find learning pattern in each of its respective attribute point's coordinates. The space represented by first layer can be expressed as,

$$H1 \in \{v0, v1, v2, ..., vn, ..., v99\} \quad (1)$$

$$\text{where } vn \in \{Xn, Yn, Zn\}$$

Subsequently, varied clusters of these neurons are fed to specific neurons in the second hidden layer. As seen from figure 1, certain clusters of attribute points (i.e. first layer neurons) represent specific parts of human face. These respective clusters can be referenced from Table 2.

Table 2: Attribute point groups for different user face regions

| User Face region | Attribute point range |
|---|---|
| Left eye | 0-7 |
| Right eye | 8-15 |
| Left eyebrow | 16-25 |
| Right eyebrow | 26-35 |
| Nose | 36-47 |
| Mouth | 48-67 |
| Face contour | 68-86 |
| Left & Right iris+nose tip | 87-89 |
| Line above left eyebrow | 90-94 |
| Line above right eyebrow | 95-99 |

Each of the second layer neurons are thus tuned to learn individual patterns in specific face regions, such as left/right eye, nose, mouth etc. respectively.

This hidden layer space can be represented as,

$$H2 \in \{H10, H11, H12, ..., H1n, ..., H19\} \quad (2)$$

$$\text{where } H1n \in \{v0 - v7, v8 - v15, v16 - v25, v26 - v35,$$
$$v36 - v47, v48 - v67, v68 - v86, v87 - v89, v90 - v94,$$
$$v95 - v99\}$$

Output layer consists of two neurons in case of binary classification task and a varying three to nine neurons in case of Multiclass classification. The second hidden layer is fully connected to each output neuron. This enables output layer neurons to fully learn patterns from each of the face regions present in H2 space. Each neuron in the architecture has an individual bias weight attached it.

## Initialization and Hyper parameter tuning

The Hyper parameters of network are optimized based upon performance comparison of different models having varying hyper parameter values. The optimized values used for model training are as shown in Table 3. `Tanh' activation function is preferred over other functions owing to its better performance for this model. `Softmax with cross entropy' function is used as activation for the output layer neurons. The weights of entire network and their biases are initialized using Xavier Initialization method [10]. Empirically, this initialization is found to perform better than random initialization for this model, in turn helping the cost function initialize closer to its global minimum.

Table 3: Model hyper parameters with optimized values

| Hyper Parameter | Optimized value |
|---|---|
| Initial learning rate | 0.01 |
| learning rate decay ratio | 0.9 |
| Rate decay step | 7000 |
| Regularization beta | 0.05 |
| Epoch | 750 |

## Network Training

The `Mean Squared Error' function is used as the cost function for this model, expressed as difference between model predictions and its true output values. For training purpose, `Adam' optimization algorithm [12] was implemented owing to its faster convergence rate, being computationally efficient and been lesser dependent on hyper parameter tuning.

The model learning rate is decreased exponentially with a decay ratio of 0.9 in every 7000 iteration steps. This implementation helps the cost function to reach its global minimum value quicker, such that it does not overshoot and miss the minimum when it is close to it, owing to larger initial learning rate. For increasing generalization capacity and to avoid over fitting of model, l2-norm regularization is used in form of a product term with hyper parameter `Regularization Beta', in order to control its contribution to the cost function.

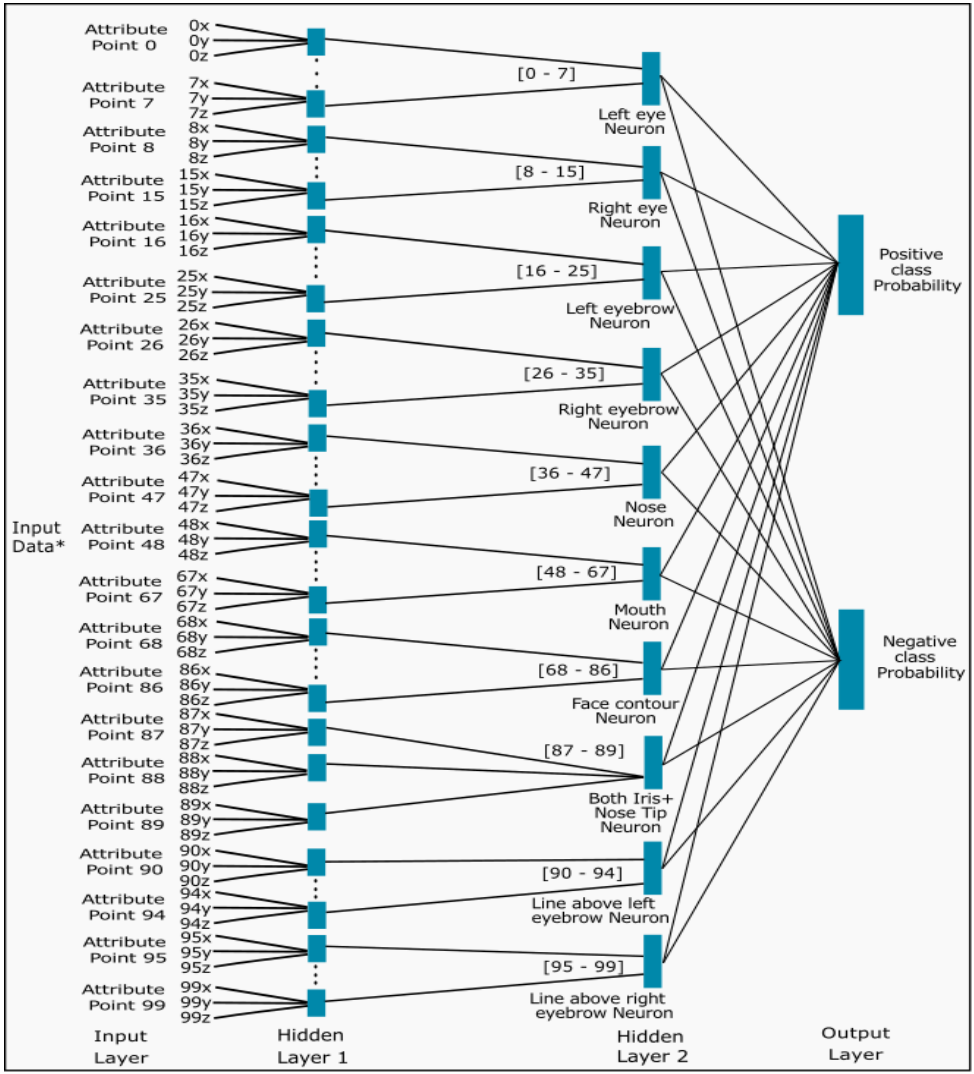

Fig. 2: Customized Deep neural network architecture

* Standardized input data using Z score method.

## Multiclass classification

For training the network to distinguish between multiple markers, four models having different number of markers to classify are implemented. These models have three, five, seven and nine markers to classify respectively. Corresponding number of neurons are present in the output layer of each model. All possible combinations of markers are used for each of these four model.

Table 4: Multiclass classification accuracy for various number of classes

| Total number of Marker classes | Percent Accuracy for test set (%) | |
| --- | --- | --- |
| | User A | User B |
| Three | 98.06 | 97.93 |
| Five | 97.79 | 97.65 |
| Seven | 97.54 | 97.31 |
| Nine | 97.12 | 96.97 |

Marker combinations are selected separately for User A and User B. Each model accuracy stated in Table 4 is calculated as a mean of these tested combinations. For training and testing the models, positive samples from each marker class in the incorporated dataset are combined to form separate data subsets. These subsets are created according to all possible marker combinations for the four models. This aggregates to about 200 - 225 samples per marker class per user.

## Results

The accuracy of proposed model on all the markers individually as a binary classification task is shown in Table 5. It also demonstrates accuracy comparison of proposed model with that of a generic fully connected network with exact same number of hidden, input and output layer neurons, keeping all its model hyper parameters and the optimization algorithm identical. The overall accuracy is calculated as mean of individual class mean accuracies over the three variants (refer Table 5), which results to 98.04%. As each marker test set differs in number of samples, better accuracy representation in terms of F score, precision and recall is shown in Table 6. These values are further compared with Freitas et al. (2014) [9]. For Multiclass classification task, accuracies of four models implemented for each user can be refer to in Table 4.

Table 5: Binary classification test set accuracy for all nine classes in comparison with those of a generic fully connected network

| Class type | Percent Accuracy for test set | | | | | |
|---|---|---|---|---|---|---|
| | Proposed model (%) | | | Fully connected network (%) | | |
| | User A | User B | Both Users | User A | User B | Both Users |
| Affirmative | 98.27 | 98.60 | 98.37 | 78.32 | 77.17 | 73.83 |
| Conditional | 97.92 | 97.86 | 97.79 | 76.42 | 76.91 | 76.39 |
| Relative | 97.34 | 97.73 | 97.49 | 82.32 | 80.63 | 81.59 |
| Negative | 98.48 | 98.34 | 98.22 | 80.45 | 79.71 | 79.18 |
| Wh Question | 97.36 | 97.83 | 97.51 | 76.82 | 75.43 | 75.71 |
| Yn Question | 98.02 | 98.49 | 98.28 | 74.51 | 74.38 | 74.41 |
| Doubt Question | 98.47 | 98.36 | 98.11 | 78.52 | 78.64 | 78.01 |
| Topics | 97.71 | 97.59 | 97.46 | 81.98 | 81.27 | 80.19 |
| Focus | 98.76 | 98.25 | 98.33 | 75.65 | 75.17 | 74.93 |
| Aggregate Mean | 98.04 | 98.12 | 97.95 | 78.33 | 77.70 | 77.14 |
| Overall Mean Accuracy | 98.04 | | | 77.72 | | |

Table 6: Binary classification Performance comparison in terms of F score, Precision and Recall

| Class Type | (Freitas et al. 2014)[a] | | | This paper | | |
|---|---|---|---|---|---|---|
| | F Score | Precision | Recall | F Score | Precision | Recall |
| Assertion | 0.89 | 0.98 | 0.90 | 0.98 | 0.97 | 0.98 |
| Conditional | 0.68 | 0.91 | 0.55 | 0.94 | 0.93 | 0.96 |
| Relative | 0.77 | 0.99 | 0.67 | 0.96 | 0.99 | 0.96 |
| Negative | 0.69 | 0.67 | 0.96 | 0.96 | 0.94 | 0.96 |
| Wh Question | 0.87 | 0.96 | 0.81 | 0.98 | 0.99 | 0.96 |
| Yn Question | 0.83 | 0.98 | 0.73 | 0.94 | 0.96 | 0.95 |
| Doubt Question | 0.89 | 0.87 | 0.94 | 0.99 | 0.97 | 0.98 |
| Topics | 0.90 | 0.95 | 0.85 | 0.98 | 0.98 | 0.97 |
| Focus | 0.91 | 0.94 | 0.89 | 0.99 | 0.98 | 0.98 |

[a] Each F-score, precision and recall value considered is the maximum taken across the four method variations in Freitas et al. 2014

## Discussion

The F-score, Precision and Recall values obtained are much improved as compared to those of accepted method present in literature. This validates that a customized network architecture as proposed here, is better capable of learning the correlation patterns in different face region coordinates for a particular expression type. The learning constraints provided to the hidden layer neurons in form of the customized architecture, help the model learn the patterns more accurately than a generic fully connected one (Table 5). For Multiclass classification, the model performs equally well and its accuracy remains mostly constant over range of different number of markers to classify.

Certain considerations include the fact that proposed model was tested on a limited dataset. This leaves it slightly untested over higher variance in certain attribute coordinates. Also this method was tested on only two users as available in the dataset. More users would have GFEs of higher degrees of variance in terms of structure of their face and method of expressing a particular GFE. A larger and more varied dataset can thus help to better train and further validate the proposed model.

## Conclusion

The overall accuracy of proposed method is excellent, such that it can reliably be used for classifying GFEs captured in form of video frames. The model performs equally well for both the Binary and Multiclass classification tasks, demonstrating its ability to correctly distinguish between a GFE and a non-GFE and between different types of GFEs. Further work on this topic would involve combining this proposed model with the current accepted methods of hand signs classification, in order to create a much evolved human sign language recognition system.

## Acknowledgements

I would like to specially thank Mr. Fernando de Almeida Freitas (Freitas, F. A.), Mr. Felipe Venâncio Barbosa (Barbosa, F. V.) and Mr. Sarajane Marques Peres (Peres, S. M.) for creating the Grammatical Facial Expressions dataset and to `University of Sao Paulo' for making it available under public license. I would also like to mention University of California, Irvine machine learning repository for hosting and maintaining this dataset.